\documentclass[onecolumn]{IEEEtran}

\usepackage{epsfig,amsmath,amssymb,amsbsy,amsthm,amsthm,scalefnt,subfig,multirow,algorithm,algorithmic,mathtools}
\usepackage{xcolor}
\usepackage{float}
\usepackage{cite}
\usepackage[T1]{fontenc}
\usepackage[doublespacing]{setspace}

\def\b0{{\pmb{0}}} 

\newcommand\ignore[1]{}

\newcommand{\indic}{\mbox{$1\!\!1$}}

\begin{document}

\title{Online Cyber-Attack Detection in Smart Grid: A Reinforcement Learning Approach}

\author{Mehmet Necip~Kurt,~%
        Oyetunji~Ogundijo,~%
        Chong~Li,~
        and Xiaodong~Wang%
\thanks{The authors are with the Department
of Electrical Engineering, Columbia University, New York, NY 10027, USA (e-mail: m.n.kurt@columbia.edu; oeo2109@columbia.edu; cl3607@columbia.edu; wangx@ee.columbia.edu).}\vspace{-2.5ex}}

\maketitle

\begin{abstract}
Early detection of cyber-attacks is crucial for a safe and reliable operation of the smart grid. In the literature, outlier detection schemes making sample-by-sample decisions and online detection schemes requiring perfect attack models have been proposed. In this paper, we formulate the online {attack/anomaly} detection problem as a partially observable Markov decision process (POMDP) problem and propose a universal robust online detection algorithm using the framework of model-free reinforcement learning (RL) for POMDPs. Numerical studies illustrate the effectiveness of the proposed RL-based algorithm in timely and accurate detection of cyber-attacks targeting the smart grid.
\end{abstract}


\section{Introduction} \label{intro}

\subsection{Background and Related Work}

The next generation power grid, i.e., the smart grid, relies on advanced control and communication technologies. This critical cyber infrastructure makes the smart grid vulnerable to hostile cyber-attacks \cite{Liang16,wang2013cyber,yan2012survey}. Main objective of attackers is to damage/mislead the state estimation mechanism in the smart grid to cause wide-area power blackouts or to manipulate electricity market prices \cite{Xie10}. There are many types of cyber-attacks, among them false data injection (FDI), jamming, and denial of service (DoS) attacks are well known. FDI attacks add malicious fake data to meter measurements \cite{Liu09,Bobba10,Li_15,Necip18}, jamming attacks corrupt meter measurements via additive noise \cite{Necip18Arxiv}, and DoS attacks block the access of system to meter measurements \cite{Asri15,zhang2011distributed,Necip18}.

The smart grid is a complex network and any failure or anomaly in a part of the system may lead to huge damages on the overall system in a short period of time. Hence, early detection of cyber-attacks is critical for a timely and effective response. In this context, the framework of quickest change detection \cite{Poor08,Basseville93,Veeravalli14,Polunchenko12} is quite useful. In the quickest change detection problems, a change occurs in the sensing environment at an unknown time and the aim is to detect the change as soon as possible with the minimal level of false alarms based on the measurements that become available sequentially over time. After obtaining measurements at a given time, decision maker either declares a change or waits for the next time interval to have further measurements. In general, as the desired detection accuracy increases, detection speed decreases. Hence, the stopping time, at which a change is declared, should be chosen to optimally balance the tradeoff between the detection speed and the detection accuracy.

If the probability density functions (pdfs) of meter measurements for the pre-change, i.e., normal system operation, and the post-change, i.e., after an attack/anomaly, cases can be modeled sufficiently accurately, the well-known cumulative sum (CUSUM) test is the optimal online detector \cite{Moustakides_86} based on the Lorden's criterion \cite{Lorden_71}. Moreover, if the pdfs can be modeled with some unknown parameters, the generalized CUSUM test, which makes use of the estimates of unknown parameters, has asymptotic optimality properties \cite{Basseville93}. However, CUSUM-based detection schemes require perfect models for both the pre- and post-change cases. In practice, capabilities of an attacker and correspondingly attack types and strategies can be totally unknown. For instance, an attacker can arbitrarily combine and launch multiple attacks simultaneously or it can launch a new unknown type of attack. Then, it may not be always possible to know the attacking strategies ahead of time and to accurately model the post-change case. Hence, universal detectors, not requiring any attack model, are needed in general. Moreover, the (generalized) CUSUM algorithm has optimality properties in minimizing a least favorable (worst-case) detection delay subject to false alarm constraints \cite{Moustakides_86,Basseville93}. Since the worst case detection delay is a pessimistic metric, it is, in general, possible to obtain algorithms performing better than the (generalized) CUSUM algorithm.


Considering the pre-change and the post-change cases as hidden states due to the unknown change-point, a quickest change detection problem can be formulated as a partially observable Markov decision process (POMDP) problem. For the problem of online attack/anomaly detection in the smart grid, in the pre-change state, the system is operated under normal conditions and using the system model, the pre-change measurement pdf can be specified highly accurately. On the other hand, the post-change measurement pdf can take different unknown forms depending on the attacker's strategy. Furthermore, the transition probability between the hidden states is unknown in general. Hence, the exact model of the POMDP is unknown.

Reinforcement learning (RL) algorithms are known to be effective in controlling uncertain environments. Hence, the described POMDP problem can be effectively solved using RL. In particular, as a solution, either the underlying POMDP model can be learned and then a model-based RL algorithm for POMDPs \cite{Ross11,Finale10} can be used or a model-free RL algorithm \cite{Jaakkola94,Perkins02,LochSingh98,Lanzi00,Peshkin99} can be used without learning the underlying model. Since the model-based approach requires a two-step solution that is computationally more demanding and only an approximate model can be learned in general, we prefer to use the model-free RL approach.

Outlier detection schemes such as the Euclidean detector \cite{Manandhar14} and the cosine-similarity metric based detector \cite{Rawat15}
are universal as they do not require any attack model. They mainly compute a dissimilarity metric between actual meter measurements and predicted measurements (by the Kalman filter) and declare an attack/anomaly if the amount of dissimilarity exceeds a certain predefined threshold. However, such detectors do not consider temporal relation between attacked/anomalous measurements and make sample-by-sample decisions. Hence, they are unable to distinguish instantaneous high-level random system noise from long-term (persistent) anomalies caused, e.g., by an unfriendly intervention to the system. Hence, compared to the outlier detection schemes, more reliable universal attack detection schemes are needed.

In this work, we consider the smart grid security problem from the defender's perspective and seek for an effective detection scheme using RL techniques {(single-agent RL)}. Note that the problem can be considered from an attacker's perspective as well, where the objective would be to determine the attacking strategies leading to the maximum possible damage on the system. Such a problem can be particularly useful in vulnerability analysis, i.e., to identify the worst possible damage an attacker may introduce to the system and accordingly to take necessary precautions. In the literature, several studies investigate vulnerability analyses using RL, see e.g., \cite{YChen18} for FDI attacks and \cite{JYan17} for sequential network topology attacks. We further note that the problem can also be considered from both defender's and attacker's perspectives simultaneously, that corresponds to a game-theoretic setting.

{Extension of single-agent RL to multiple agents is the multi-agent RL framework that quite involves game theory since in this case, the optimal policies of agents depend both on the environment and the policies of the other agents. Moreover, stochastic games extend the Markov decision processes to multi-agent case where the game is sequential and consists of multiple states, and the transition from one state to another and also the payoffs (rewards/costs) depend on joint actions of all agents. Several RL-based solution approaches have been proposed for stochastic games, see e.g., \cite{nowe2012,littman1994,claus1998,hu2003nash,weinberg2004}. Further, if the game (the underlying state of the environment, actions and payoffs of other agents, etc.) is partially observed, then it is called a partially observable stochastic game, for which finding a solution is more difficult in general.}

\subsection{Contributions}

In this paper, we propose an online cyber-attack detection algorithm using the framework of model-free RL for POMDPs. The proposed  algorithm is universal, i.e., it does not require attack models. This makes the proposed scheme widely applicable and also proactive in the sense that new unknown attack types can be detected. Since we follow a model-free RL approach, the defender learns a direct mapping from observations to actions (\emph{stop} or \emph{continue}) by trial-and-error. In the training phase, although it is possible to obtain/generate observation data for the pre-change case using the system model under normal operating conditions, it is generally difficult to obtain real attack data. For this reason, we follow a robust detection approach by training the defender with low-magnitude attacks that corresponds to the worst-case scenarios from a defender's perspective since such attacks are quite difficult to detect. Then, the trained defender becomes sensitive to detect slight deviations of meter measurements from the normal system operation. The robust detection approach significantly limits the action space of an attacker as well. That is, to prevent the detection, an attacker can only exploit very low attack magnitudes that are practically not much of interest due to their minimal damage on the system. To the best of our knowledge, this work is the first attempt for online cyber-attack detection in the smart grid using RL techniques.

\subsection{Organization and Notation}

We introduce the system model and the state estimation mechanism in Sec.~\ref{sys_mod}. We present the problem formulation in Sec.~\ref{prob_form} and the proposed solution approach in Sec.~\ref{solutions}. We then illustrate the performance of the proposed RL-based detection scheme via extensive simulations in Sec.~\ref{numerical}. Finally, we conclude the paper in Sec.~\ref{conclusion}. Boldface letters denote vectors and matrices, all vectors are column vectors, and $\pmb{o}^\mathrm{T}$ denotes the transpose of $\pmb{o}$. $\mathrm{P}$ and $\mathrm{E}$ denote the probability and expectation operators, respectively. Table~\ref{table:symbols} summarizes the common symbols and parameters used in the paper.

\renewcommand{\arraystretch}{1.1} 
\begin{table}[t]

    \centering
    \begin{tabular}{ | p{1cm} | l |}
    \hline
    Symbol & Meaning  \\ \hline \hline
    $\Gamma$ & Stopping time  \\ \hline
    $\tau$ & Change-point  \\ \hline
    $I$ & Number of quantization levels  \\ \hline
    $M$ & Window size  \\ \hline
    $Q(o,a)$ & $Q$-value corresponding to observation-action pair $(o,a)$ \\ \hline
    $\alpha$ & Learning rate  \\ \hline
    $\epsilon$ & Exploration rate  \\ \hline
    $T$ & Maximum length of a learning episode  \\ \hline
    $E$ & Number of learning episodes  \\
    \hline
    \end{tabular}
    \caption{{Common symbols/parameters in the paper.}}
    \label{table:symbols}
\end{table}
\renewcommand{\arraystretch}{1} 

\section{System Model and State Estimation} \label{sys_mod}

\subsection{System Model}

Suppose that there are $K$ meters in a power grid consisting of $N+1$ buses, where usually $K > N$ to have the necessary measurement redundancy against noise \cite{Abur04}. One of the buses is considered as a reference bus and the system state at time $t$ is denoted with $\mathbf{x}_t = [x_{1,t}, \dots, x_{N,t}]^\mathrm{T}$ where $x_{n,t}$ denotes the phase angle at bus $n$ at time $t$. Let the measurement taken at meter $k$ at time $t$ be denoted with $y_{k,t}$ and the measurement vector be denoted with $\mathbf{y}_t = [y_{1,t}, \dots, y_{K,t}]^\mathrm{T}$. Based on the widely used linear DC model \cite{Abur04}, we model the smart grid with the following state-space equations:
\begin{gather} \label{eq:state_upd}
\mathbf{x}_{t} = \mathbf{A} \mathbf{x}_{t-1} + \mathbf{v}_t, \\ \label{eq:meas_model}
\mathbf{y}_t = \mathbf{H} \mathbf{x}_t + \mathbf{w}_t,
\end{gather}
where $\mathbf{A} \in \mathbb{R}^{N \times N}$ is the system (state transition) matrix, $\mathbf{H} \in \mathbb{R}^{K \times N}$ is the measurement matrix determined based on the network topology, $\mathbf{v}_t = [v_{1,t}, \dots, v_{N,t}]^\mathrm{T}$ is the process noise vector, and ${\mathbf{w}_t = [w_{1,t}, \dots, w_{K,t}]^\mathrm{T}}$ is the measurement noise vector. We assume that $\mathbf{v}_t$ and $\mathbf{w}_t$ are independent additive white Gaussian random processes where ${\mathbf{v}_t \sim \mathbf{\mathcal{N}}(\mathbf{0},\sigma_v^2 \, \mathbf{I}_N)}$, ${\mathbf{w}_t \sim \mathbf{\mathcal{N}}(\mathbf{0},\sigma_w^2 \, \mathbf{I}_K)}$, and $\mathbf{I}_K \in \mathbb{R}^{K \times K}$ is an identity matrix. Moreover, we assume that the system is observable, i.e., the observability matrix
\begin{equation}\nonumber
\mathbf{O} \triangleq \left[
  \begin{smallmatrix}
    \mathbf{H} \\
    \mathbf{H} \mathbf{A} \\
    \vdots \\
    \mathbf{H} \mathbf{A}^{N-1}
  \end{smallmatrix}
\right]
\end{equation}
has rank $N$.

The system model given in \eqref{eq:state_upd} and \eqref{eq:meas_model} corresponds to the normal system operation. In case of a cyber-attack, however, the measurement model in \eqref{eq:meas_model} is no longer true. For instance,
\begin{enumerate}
  \item in case of an FDI attack launched at time $\tau$, the measurement model can be written as
      \begin{equation}\nonumber
        \mathbf{y}_t = \mathbf{H} \mathbf{x}_t + \mathbf{w}_t + \mathbf{b}_t \indic \{t \geq \tau\},
      \end{equation}
      where $\indic$ is an indicator function and $\mathbf{b}_t \triangleq [b_{1,t},\dots,b_{K,t}]^\mathrm{T}$ denotes the injected malicious data at time $t \geq \tau$ {and $b_{k,t}$ denotes the injected false datum to the $k$th meter at time $t$},
  \item in case of a jamming attack with additive noise, the measurement model can be written as
      \begin{equation}\nonumber
        \mathbf{y}_t = \mathbf{H} \mathbf{x}_t + \mathbf{w}_t + \mathbf{u}_t \indic \{t \geq \tau\},
      \end{equation}
      where $\mathbf{u}_t \triangleq [u_{1,t},\dots,u_{K,t}]^\mathrm{T}$ denotes the random noise realization at time $t \geq \tau$ {and $u_{k,t}$ denotes the jamming noise corrupting the $k$th meter at time $t$},
  \item {in case of a hybrid FDI/jamming attack \cite{Necip18Arxiv}, the meter measurements take the following form:
       \begin{equation}\nonumber
        \mathbf{y}_t = \mathbf{H} \mathbf{x}_t + \mathbf{w}_t + (\mathbf{b}_t + \mathbf{u}_t) \indic \{t \geq \tau\},
      \end{equation}}
  \item in case of a DoS attack, meter measurements can be partially unavailable to the system controller. The measurement model can then be written as
      \begin{equation}\nonumber
        \mathbf{y}_t = \mathbf{D}_t (\mathbf{H} \mathbf{x}_t + \mathbf{w}_t),
      \end{equation}
      where $\mathbf{D}_t = \mathrm{diag}(d_{1,t}, \dots, d_{K,t})$ is a diagonal matrix consisting of $0$s and $1$s. Particularly, if $y_{k,t}$ is available, then $d_{k,t} = 1$, otherwise $d_{k,t} = 0$. Note that $\mathbf{D}_t = \mathbf{I}_K$ for $t<\tau$,
  \item {in case of a network topology attack, the measurement matrix changes. Denoting the measurement matrix under topology attack at time $t\geq\tau$ by $\bar{\mathbf{H}}_t$, we have
      \begin{equation}\nonumber
        \mathbf{y}_t =
        \begin{cases}
          \mathbf{H} \mathbf{x}_t + \mathbf{w}_t, & \mbox{if } t<\tau \\
          \bar{\mathbf{H}}_t \mathbf{x}_t + \mathbf{w}_t, & \mbox{if } t \geq \tau,
        \end{cases}
      \end{equation}}
  \item {in case of a mixed topology and hybrid FDI/jamming attack, the measurement model can be written as follows:
      \begin{equation}\nonumber
        \mathbf{y}_t =
        \begin{cases}
          \mathbf{H} \mathbf{x}_t + \mathbf{w}_t, & \mbox{if } t<\tau \\
          \bar{\mathbf{H}}_t \mathbf{x}_t + \mathbf{w}_t + \mathbf{b}_t + \mathbf{u}_t, & \mbox{if } t \geq \tau.
        \end{cases}
      \end{equation}}
\end{enumerate}

\subsection{State Estimation}

Since the smart grid is regulated based on estimated system states, state estimation is a fundamental task in the smart grid, that is conventionally performed using the static least squares (LS) estimators \cite{Liu09,Bobba10,Esmalifalak11}. However, in practice, the smart grid is a highly dynamic system due to time-varying load and power generation \cite{Tan17}. Furthermore, time-varying cyber-attacks can be designed and performed by the adversaries. Hence, dynamic system modeling as in \eqref{eq:state_upd} and \eqref{eq:meas_model} and correspondingly using a dynamic state estimator can be quite useful for real-time operation and security of the smart grid \cite{Necip18,Necip18Arxiv}.

For a discrete-time linear dynamic system, if the noise terms are Gaussian, the Kalman filter is the optimal linear estimator in minimizing the mean squared state estimation error \cite{Kalman_60}. Note that for the Kalman filter to work correctly, the system needs to be observable. The Kalman filter is an online estimator consisting of prediction and measurement update steps at each iteration. Denoting the state estimates at time $t$ with $\hat{\mathbf{x}}_{t|t'}$ where $t' = t-1$ and $t' = t$ for the prediction and measurement update steps, respectively, the Kalman filter equations at time $t$ can be written as follows:

\emph{Prediction}:
\begin{gather} \nonumber
\hat{\mathbf{x}}_{t|t-1} = \mathbf{A} \hat{\mathbf{x}}_{t-1|t-1}, \\ \label{eq:pred}
\mathbf{F}_{t|t-1} = \mathbf{A} \mathbf{F}_{t-1|t-1} \mathbf{A}^\mathrm{T} + \sigma_v^2 \, \mathbf{I}_N,
\end{gather}
\emph{Measurement update}:
\begin{gather} \nonumber
\mathbf{G}_{t} = \mathbf{F}_{t|t-1} \mathbf{H}^\mathrm{T} (\mathbf{H} \mathbf{F}_{t|t-1} \mathbf{H}^\mathrm{T} + \sigma_w^2 \, \mathbf{I}_K)^{-1}, \\ \nonumber
\hat{\mathbf{x}}_{t|t} = \hat{\mathbf{x}}_{t|t-1} + \mathbf{G}_{t} (\mathbf{y}_t - \mathbf{H} \hat{\mathbf{x}}_{t|t-1}), \\ \label{eq:meas_upd_fdata}
\mathbf{F}_{t|t} = \mathbf{F}_{t|t-1} - \mathbf{G}_{t} \mathbf{H} \mathbf{F}_{t|t-1},
\end{gather}
where $\mathbf{F}_{t|t-1}$ and $\mathbf{F}_{t|t}$  denote the estimates of the state covariance matrix based on the measurements up to $t-1$ and $t$, respectively. Moreover, $\mathbf{G}_{t}$ is the Kalman gain matrix at time $t$.

\begin{figure}[t]
\center
  \includegraphics[width=77mm]{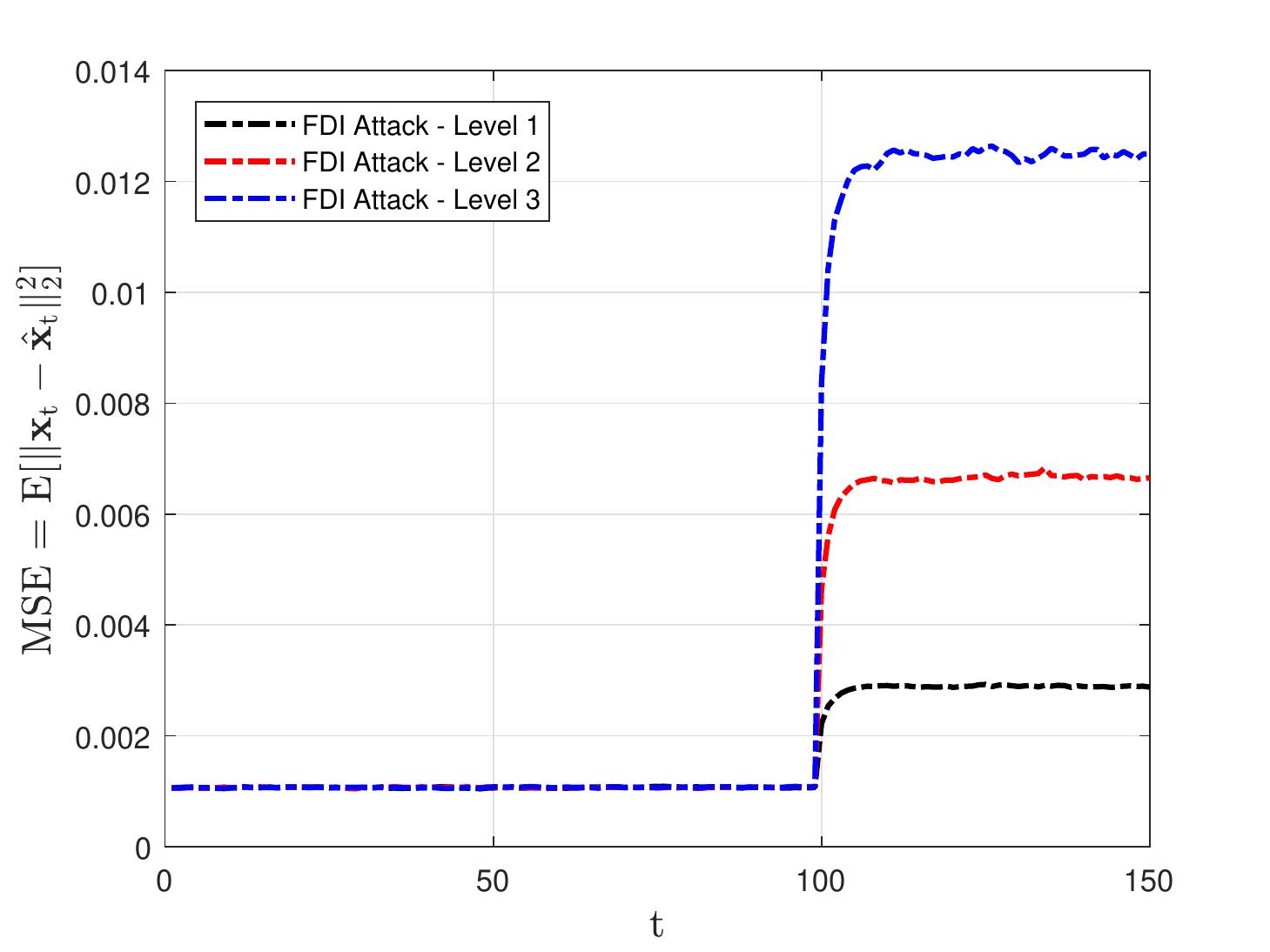}
\caption{{Mean squared state estimation error vs. time where random FDI attacks with various magnitude levels are launched at time $\tau = 100$.}}
 \label{fig:MSE_vs_Time}
\end{figure}

{We next demonstrate the effect of cyber-attacks on the state estimation mechanism via an illustrative example. We consider a random FDI attack with various magnitude/intensity levels and show how the mean squared state estimation error of the Kalman filter changes when FDI attacks are launched to the system. We assume that the attacks are launched at $\tau = 100$, i.e., the system is operated under normal (non-anomalous) conditions up to time $100$ and under attacking conditions afterwards. Three attack magnitude levels are considered:
\begin{itemize}
  \item Level 1: $b_{k,t} \sim \mathcal{U}[-0.04,0.04]$, $\forall k \in \{1,\dots,K\}$, $\forall t \geq \tau$,
  \item Level 2: $b_{k,t} \sim \mathcal{U}[-0.07,0.07]$, $\forall k \in \{1,\dots,K\}$, $\forall t \geq \tau$,
  \item Level 3: $b_{k,t} \sim \mathcal{U}[-0.1,0.1]$, $\forall k \in \{1,\dots,K\}$, $\forall t \geq \tau$,
\end{itemize}
where $\mathcal{U}[\zeta_1,\zeta_2]$ denotes a uniform random variable in the range of $[\zeta_1,\zeta_2]$. The corresponding mean squared error (MSE) versus time curves are presented in Fig.~\ref{fig:MSE_vs_Time}. We observe that in case of cyber-attacks, the state estimates are deviated from the actual system states where the amount of deviation increases as the attack magnitudes get larger.}

\section{Problem Formulation} \label{prob_form}

Before we introduce our problem formulation, we briefly explain a POMDP setting as follows. Given an agent and an environment, a discrete-time POMDP is defined by the seven-tuple $(\mathcal{S}, \mathcal{A}, \mathcal{T}, \mathcal{R}, \mathcal{O},  \mathcal{G}, \gamma)$ where $\mathcal{S}$ denotes the set of (hidden) states of the environment, $\mathcal{A}$ denotes the set of actions of the agent, $\mathcal{T}$ denotes the set of conditional transition probabilities between the states, $\mathcal{R}: \mathcal{S} \times \mathcal{A} \rightarrow \mathbb{R}$ denotes the reward function that maps the state-action pairs to rewards, $\mathcal{O}$ denotes the set of observations of the agent, $\mathcal{G}$ denotes the set of conditional observation probabilities, and $\gamma \in [0, 1]$ denotes a discount factor that indicates how much present rewards are preferred over the future rewards.

At each time $t$, the environment is in a particular hidden state $s_t \in \mathcal{S}$. Obtaining an observation $o_t \in \mathcal{O}$ depending on the current state of the environment with the probability $\mathcal{G}(o_t|s_t)$, the agent takes an action $a_t \in \mathcal{A}$ and receives a reward $r_t = \mathcal{R}(s_t,a_t)$ from the environment based on its action and the current state of the environment. At the same time, the environment makes a transition to the next state $s_{t+1}$ with the probability $\mathcal{T}(s_{t+1}|s_t,a_t)$. The process is repeated until a terminal state is reached. In this process, the goal of the agent is to determine an optimal policy $\pi: \mathcal{O} \rightarrow \mathcal{A}$ that maps observations to actions and maximizes the agent's expected total discounted reward, i.e., $\mathrm{E} \big[  \sum_{t = 0}^{\infty} \gamma^{t} r_{t} \big]$. Equivalently, if an agent receives costs instead of rewards from the environment, then the goal is to minimize the expected total discounted cost. Considering the latter, the POMDP problem can be written as follows:
\begin{equation}\label{eq:POMDP}
\min_{\pi: \, \mathcal{O} \rightarrow \mathcal{A}}~ \mathrm{E} \Big[ \sum_{t = 0}^{\infty} \gamma^{t} r_{t} \Big].
\end{equation}

\begin{figure}[t]
\vspace{-0.4cm}
\center
  \includegraphics[width=64mm]{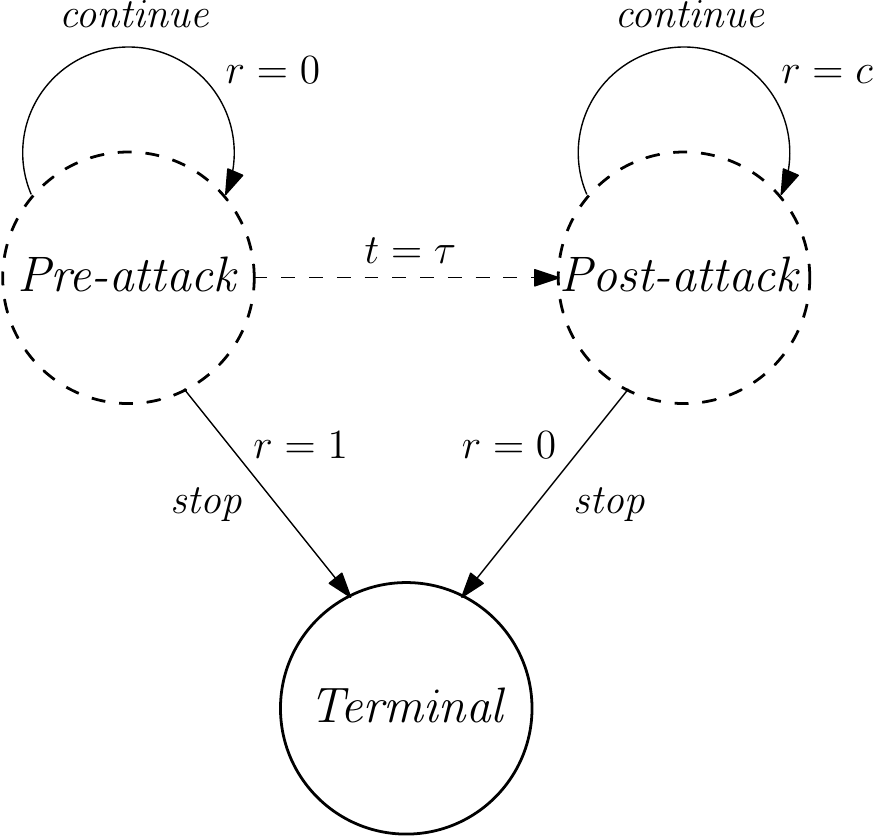}
\caption{State-machine diagram for the considered POMDP setting. The hidden states and the (hidden) transition between them happening at time $t = \tau$ are illustrated with the dashed circles and the dashed line, respectively. The defender receives costs ($r$) depending on its actions and the underlying state of the environment. Whenever the defender chooses the action \emph{stop}, the system moves into a \emph{terminal} state and the defender receives no further cost.}
 \label{fig:state_machine}
\end{figure}

Next, we explain the online attack detection problem in a POMDP setting. We assume that at an unknown time $\tau$, a cyber-attack is launched to the system and our aim is to detect the attack as quickly as possible after it occurs, where the attacker's capabilities/strategies are completely unknown. This defines a quickest change detection problem where the aim is to minimize the average detection delay as well as the false alarm rate. This problem can, in fact, be expressed as a POMDP problem (see Fig.~\ref{fig:state_machine}). In particular, due to the unknown attack launch time $\tau$, there are two hidden states: \emph{pre-attack} and \emph{post-attack}. At each time $t$, after obtaining the measurement vector $\mathbf{y}_t$, two actions are available for the agent (defender): \emph{stop} and declare an attack or \emph{continue} to have further measurements. We assume that whenever the action \emph{stop} is chosen, the system moves into a \emph{terminal} state, and always stays there afterwards.

Furthermore, although the conditional observation probability for the \emph{pre-attack} state can be inferred based on the system model under normal operating conditions, since the attacking strategies are unknown, the conditional observation probability for the \emph{post-attack} state is assumed to be totally unknown. Moreover, due to the unknown attack launch time $\tau$, state transition probability between the \emph{pre-attack} and the \emph{post-attack} states is unknown.

Since our aim is to minimize the detection delays and the false alarm rate, both the false alarm and the detection delay events should be associated with some costs. Let the relative cost of a detection delay compared to a false alarm event be $c>0$. Then, if the true underlying state is \emph{pre-attack} and the action \emph{stop} is chosen, a false alarm occurs and the defender receives a cost of $1$. On the other hand, if the underlying state is \emph{post-attack} and the action \emph{continue} is chosen, then the defender receives a cost of $c$ due to the detection delay. For all other (hidden) state-action pairs, the cost is assumed to be zero. Also, once the action \emph{stop} is chosen, the defender does not receive any further costs while staying in the \emph{terminal} state. The objective of the defender is to minimize its expected total cost by properly choosing its actions. Particularly, based on its observations, the defender needs to determine the stopping time at which an attack is declared.

Let $\Gamma$ denote the stopping time chosen by the defender. Moreover, let $\mathrm{P}_k$ denote the probability measure if the attack is launched at time $k$, i.e., $\tau = k$, and let $\mathrm{E}_k$ denote the corresponding expectation. Note that since the attacking strategies are unknown, $\mathrm{P}_k$ is assumed to be unknown. For the considered online attack detection problem, we can derive the expected total discounted cost as follows:
\begin{align} \nonumber
\mathrm{E} \Big[ \sum_{t = 0}^{\infty} \gamma^{t} r_{t} \Big] &= \mathrm{E}_\tau \Big[ \indic\{\Gamma < \tau\} + \sum_{t = \tau}^{\Gamma} c \Big] \\ \nonumber
&= \mathrm{E}_\tau \big[ \indic\{\Gamma < \tau\} + c \, (\Gamma-\tau)^+ \big] \\ \label{eq:obj_func}
&= \mathrm{P}_\tau (\{\Gamma < \tau\}) + c \, \mathrm{E}_\tau \big[(\Gamma-\tau)^+\big],
\end{align}
where $\gamma = 1$ is chosen since the present and future costs are equally weighted in our problem, $\{\Gamma < \tau\}$ is a false alarm event that is penalized with a cost of $1$, and $\mathrm{E}_\tau \big[(\Gamma-\tau)^+\big]$ is the average detection delay where each detection delay is penalized with a cost of $c$ and $(\cdot)^+ = \max(\cdot, 0)$.

Based on \eqref{eq:POMDP} and \eqref{eq:obj_func}, the online attack detection problem can be written as follows:
\begin{gather} \label{eq:opt_prob1}
\min_{\Gamma}~ \mathrm{P}_\tau (\{\Gamma < \tau\}) + c \, \mathrm{E}_\tau \big[(\Gamma-\tau)^+\big].
\end{gather}
Since $c$ corresponds to the relative cost between the false alarm and the detection delay events, by varying $c$ and solving the corresponding problem in \eqref{eq:opt_prob1}, a tradeoff curve between average detection delay and false alarm rate can be obtained. Moreover, $c < 1$ can be chosen to prevent frequent false alarms.

Since the exact POMDP model is unknown due to unknown attack launch time $\tau$ and the unknown attacking strategies and since the RL algorithms are known to be effective over uncertain environments, we follow a model-free RL approach to obtain a solution to \eqref{eq:opt_prob1}. Then, a direct mapping from observations to the actions, i.e., the stopping time $\Gamma$, needs to be learned. Note that the optimal action is \emph{continue} if the underlying state is \emph{pre-attack} and \emph{stop} if the underlying state is \emph{post-attack}. Then, to determine the optimal actions, the underlying state needs to be inferred using observations and the observation signal should be well informative to reduce the uncertainty about the underlying state. As described in Sec.~\ref{sys_mod}, the defender observes the measurements $\mathbf{y}_t$ at each time $t$. The simplest approach can be forming the observation space directly with the measurement vector $\mathbf{y}_t$ but we would like to process the measurements and form the observation space with a signal related to the deviation of system from its normal operation.

Furthermore, it is, in general, possible to obtain identical observations in the \emph{pre-attack} and the \emph{post-attack} states. This is called perceptual aliasing and prevent us to make a good inference about the underlying state by only looking at the observation at a single time. We further note that in our problem, deciding on an attack solely based on a single observation corresponds to an outlier detection scheme for which more practical detectors are available not requiring a learning phase, see e.g., \cite{Manandhar14,Rawat15}. However, we are particularly interested in detecting sudden and persistent attacks/anomalies that more likely happen due to an unfriendly intervention to the system rather than random disturbances due to high-level system noise realizations.

Since different states require different optimal actions, the ambiguity on the underlying state should be further reduced with additional information derived from the history of observations. In fact, there may be cases where the entire history of observations is needed to determine the optimal solution in a POMDP problem \cite{Meuleau97}. However, due to computational limitations, only a finite memory can be used in practice and an approximately optimal solution can be obtained. A simple approach is to use a finite-size sliding window of observations as a memory and map the most recent history window to an action, as described in \cite{LochSingh98}. This approach is particularly suitable for our problem as well since we assume persistent attacks/anomalies that happen at an unknown point of time and continue thereafter. That is, only the observations obtained after an attack are significant from the attack detection perspective.

\begin{figure}[t]
\vspace{-0.4cm}
\center
  \includegraphics[width=88mm]{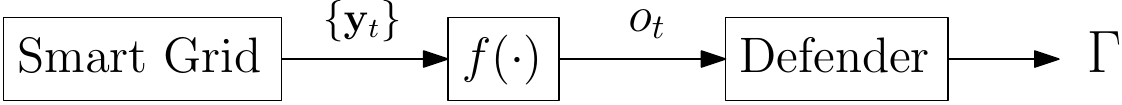}
\caption{A graphical description of the online attack detection problem in the smart grid. The measurements $\{\mathbf{y}_t\}$ are collected through smart meters and processed to obtain $o_t = f(\{\mathbf{y}_t\})$. The defender observes $f(\{\mathbf{y}_t\})$ at each time $t$ and decides on the attack declaration time $\Gamma$.}
 \label{fig:prob_form}
\end{figure}

Let the function that processes a finite history of measurements and produces the observation signal be denoted with $f(\cdot)$ so that the observation signal at time $t$ is $o_t = f(\{\mathbf{y}_t\})$. Then, at each time, the defender observes $f(\{\mathbf{y}_t\})$ and decides on the stopping time $\Gamma$, as illustrated in Fig.~\ref{fig:prob_form}. The aim of the defender is to obtain a solution to \eqref{eq:opt_prob1} by using an RL algorithm, as detailed in the subsequent section.

\section{Solution Approach} \label{solutions}

Firstly, we explain our methodology to obtain the observation signal $o_t = f(\{\mathbf{y}_t\})$. Note that the pdf of meter measurements in the \emph{pre-attack} state can be inferred using the baseline measurement model in \eqref{eq:meas_model} and the state estimates provided by the Kalman filter. In particular, the pdf of the measurements under normal operating conditions can be estimated as follows:
\begin{gather} \nonumber
\mathbf{y}_t \sim \mathbf{\mathcal{N}}(\mathbf{H} \hat{\mathbf{x}}_{t|t},\sigma_w^2 \, \mathbf{I}_K).
\end{gather}
The likelihood of measurements based on the baseline density estimate, denoted with $L(\mathbf{y}_t)$, can then be computed as follows:
\begin{align} \nonumber
L(\mathbf{y}_t) &= (2 \pi \sigma_w^2)^{-\frac{K}{2}} \exp \Big(\frac{-1}{2 \sigma_w^2} (\mathbf{y}_t - \mathbf{H} \hat{\mathbf{x}}_{t|t})^\mathrm{T} (\mathbf{y}_t - \mathbf{H} \hat{\mathbf{x}}_{t|t}) \Big) \\ \nonumber
&= (2 \pi \sigma_w^2)^{-\frac{K}{2}} \exp \Big(\frac{-1}{2 \sigma_w^2} \eta_t \Big),
\end{align}
where
\begin{gather} \label{eq:eta_t}
\eta_t \triangleq (\mathbf{y}_t - \mathbf{H} \hat{\mathbf{x}}_{t|t})^\mathrm{T} (\mathbf{y}_t - \mathbf{H} \hat{\mathbf{x}}_{t|t})
\end{gather}
is the estimate of the negative log-scaled likelihood.

In case the system is operated under normal conditions, the likelihood $L(\mathbf{y}_t)$ is expected to be high. Equivalently, small (close to zero) values of $\eta_t$ may indicate the normal system operation. On the other hand, in case of an attack/anomaly, the system deviates from normal operating conditions and hence the likelihood $L(\mathbf{y}_t)$ is expected to decrease in such cases. Then, persistent high values of $\eta_t$ over a time period may indicate an attack/anomaly. Hence, $\eta_t$ may help to reduce the uncertainty about the underlying state to some extent. 

However, since $\eta_t$ can take any nonnegative value, the observation space is continuous and hence learning a mapping from each possible observation to an action is computationally infeasible. To reduce the computational complexity in such continuous spaces, we can quantize the observations. We then partition the observation space into $I$ mutually exclusive and disjoint intervals using the quantization thresholds $\beta_0 = 0 < \beta_1 < \dots < \beta_{I-1} < \beta_I = \infty$ so that if $\beta_{i-1} \leq \eta_t < \beta_{i}, i \in {1,\dots,I}$, the observation at time $t$ is represented with $\theta_i$. Then, possible observations at any given time are $\theta_1, \dots, \theta_I$. Since $\theta_i$'s are representations of the quantization levels, each $\theta_i$ needs to be assigned to a different value.

Furthermore, as explained before, although $\eta_t$ may be useful to infer the underlying state at time $t$, it is possible to obtain identical observations in the \emph{pre-attack} and \emph{post-attack} states. For this reason, we propose to use a finite history of observations. Let the size of the sliding observation window be $M$ so that there are $I^M$ possible observation windows and the sliding window at time $t$ consists of the quantized versions of ${\{\eta_j: t-M+1 \leq j \leq t\}}$. Henceforth, by an observation $o$, we refer to an observation window so that the observation space $\mathcal{O}$ consists of all possible observation windows. For instance, if $I = M = 2$, then $\mathcal{O} = \{ [\theta_1, \theta_1], [\theta_1, \theta_2], [\theta_2, \theta_1], [\theta_2, \theta_2] \}$.

For each possible observation-action pair $(o,a)$, we propose to learn a $Q(o,a)$ value, i.e., the expected future cost, using an RL algorithm where all $Q(o,a)$ values are stored in a $Q$-table of size $I^M \times 2$. After learning the $Q$-table, the policy of the defender will be choosing the action $a$ with the minimum $Q(o,a)$ for each observation $o$. In general, increasing $I$ and $M$ may improve the learning performance but at the same time results in a larger $Q$ table, that would require to increase the number of training episodes and hence the computational complexity of the learning phase. Hence, $I$ and $M$ should be chosen considering the expected tradeoff between performance and computational complexity.

\begin{algorithm}[t]\small
\caption{\small Learning Phase -- SARSA Algorithm}
\label{alg:training}
\baselineskip=0.37cm
\begin{algorithmic}[1]
\STATE Initialize $Q(o,a)$ arbitrarily, $\forall o \in \mathcal{O}$ and $\forall a \in \mathcal{A}$.
\FOR {$e = 1:E$}
    \STATE $t \gets 0$
    \STATE $s \gets$ \emph{pre-attack}
    \STATE Choose an initial $o$ based on the \emph{pre-attack} state and choose the initial $a = \emph{continue}$.
    \WHILE {$s \neq$ \emph{terminal} and $t < T$}
        \STATE $t \gets t+1$
        \IF {$a =$ \emph{stop}}
            \STATE $s \gets$ \emph{terminal}
            \STATE $r \gets \indic\{t < \tau\}$
            \STATE $Q(o,a) \gets Q(o,a) + \alpha \, (r - Q(o,a))$
        \ELSIF {$a =$ \emph{continue}}
            \IF {$t >= \tau$}
                \STATE $r \gets c$
                \STATE $s \gets$ \emph{post-attack}
            \ELSE
                \STATE $r \gets 0$
            \ENDIF
            \STATE Collect the measurements $\mathbf{y}_t$.
            \STATE Employ the Kalman filter using \eqref{eq:pred} and \eqref{eq:meas_upd_fdata}.
            \STATE Compute $\eta_t$ using \eqref{eq:eta_t} and quantize it to obtain $\theta_i$ if $\beta_{i-1} \leq \eta_t < \beta_{i}, i \in {1,\dots,I}$.
            \STATE Update the sliding observation window $o$ with the most recent entry $\theta_i$ and obtain $o'$.
            \STATE Choose action $a'$ from $o'$ using the $\epsilon$-greedy policy based on the $Q$-table (that is being learned).
            \STATE $Q(o,a) \gets Q(o,a) + \alpha \, (r + Q(o',a') - Q(o,a))$
            \STATE $o \gets o'$, $a \gets a'$
        \ENDIF
    \ENDWHILE
\ENDFOR
\STATE Output: $Q$-table, i.e., $Q(o,a)$, $\forall o \in \mathcal{O}$ and $\forall a \in \mathcal{A}$.
\end{algorithmic}
\end{algorithm}

\begin{algorithm}[t]\small
\caption{\small Online Attack Detection}
\label{alg:test}
\baselineskip=0.37cm
\begin{algorithmic}[1]
\STATE Input: $Q$-table learned in Algorithm \ref{alg:training}.
\STATE Choose an initial $o$ based on the \emph{pre-attack} state and choose the initial $a = \emph{continue}$.
\STATE $t \gets 0$
\WHILE {$a \neq$ \emph{stop}}
    \STATE $t \gets t+1$
    \STATE Collect the measurements $\mathbf{y}_t$.
    \STATE Determine the new $o$ as in the lines 20--22 of Algorithm 1.
    \STATE $a \gets {\arg \min}_{a} Q(o,a)$.
\ENDWHILE
\STATE Declare an attack and terminate the procedure.
\end{algorithmic}
\end{algorithm}

\begin{figure}[t]
\vspace{-0.4cm}
\center
  \includegraphics[width=64mm]{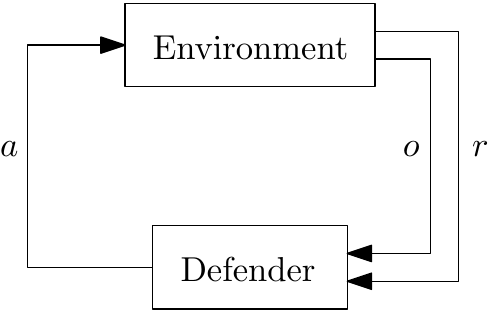}
\caption{An illustration of the interaction between defender and the simulation environment during the learning procedure. The environment provides an observation $o$ based on its internal state $s$, the agent chooses an action $a$ based on its observation and receives a cost $r$ from the environment in return of its action. Based on this experience, the defender updates $Q(o,a)$. This process is repeated many times during the learning procedure.}
 \label{fig:RL}
\end{figure}

The considered RL-based detection scheme consists of learning and online detection phases. In the literature, SARSA, a model-free RL control algorithm \cite{Sutton98reinforcement}, was numerically shown to perform well over the model-free POMDP settings \cite{Peshkin99}. Hence, in the learning phase, the defender is trained with many episodes of experience using the SARSA algorithm and a $Q$-table is learned by the defender. For training, a simulation environment is created and during the training procedure, at each time, the defender takes an action based on its observation and receives a cost in return of its action from the simulation environment, as illustrated in Fig.~\ref{fig:RL}. Based on this experience, the defender updates and learns a $Q$-table. Then, in the online detection phase, based on the observations, the action with the lowest expected future cost ($Q$ value) is chosen at each time using the previously learned $Q$-table. The online detection phase continues until the action \emph{stop} is chosen by the defender. Whenever \emph{stop} is chosen, an attack is declared and the process is terminated.

Note that after declaring an attack, whenever the system is recovered and returned back to the normal operating conditions, the online detection phase can be restarted. That is, once a defender is trained, no further training is needed. We summarize the learning and the online detection stages in Algorithms \ref{alg:training} and \ref{alg:test}, respectively. In Algorithm~\ref{alg:training}, $E$ denotes the number of learning episodes, $T$ denotes the maximum length of a learning episode, $\alpha$ is the learning rate, and $\epsilon$ is the exploration rate, where the $\epsilon$-greedy policy chooses the action with the minimum $Q$ value with probability $1-\epsilon$ and the other action (for exploration purposes during the learning process) with probability $\epsilon$.

{Since RL is an iterative procedure, same actions are repeated at each iteration (learning episode). The time complexity of an RL algorithm can then be considered as the time complexity of a single iteration \cite{Kokar}. As the SARSA algorithm performs one update on the $Q$-table at a time and the maximum time limit for a learning episode is $T$, the time complexity of Algorithm 1 is $O(T)$. Moreover, the overall complexity of the learning procedure is $O(T E)$, as $E$ is the number of learning episodes. Notice that the time complexity does not depend on the size of the action and observation spaces. On the other hand, as $I$ and/or $M$ increase, a larger $Q$-table needs to be learned, that requires to increase $E$ for a better learning. Furthermore, the space complexity (memory cost) of Algorithm 1 is $M + 2 \, I^M$ due to the sliding observation window of size $M$ and the $Q$-table of size $I^M\times2$. Note that the space complexity is fixed over time. During the learning procedure, based on the smart grid model and some attack models (that are used to obtain low-magnitude attacks that correspond to small deviations from the normal system operation), we can obtain the measurement data online and the defender is trained with the observed data stream. Hence, the learning phase does not require storage of large amount of training data as only a sliding observation window of size $M$ needs to be stored at each time.}

{In Algorithm 2, at each time, the observation $o$ is determined and using the $Q$-table (learned in Algorithm 1), the corresponding action $a$ with the minimum cost is chosen. Hence, the complexity at a time is $O(1)$. This process is repeated until the action \emph{stop} is chosen at the stopping time $\Gamma$. Furthermore, similarly to Algorithm 1, the space complexity of Algorithm 2 is $M + 2 \, I^M$.}

{\textit{Remark 1:} Since our solution approach is model-free, i.e., it is not particularly designed for specific types of attacks, the proposed detector does not distinguish between an attack and other types of persistent anomalies such as network topology faults. In fact, the proposed algorithm can detect any attack/anomaly as long as effect of such attack/anomaly on the system is at a distinguishable level, i.e., the estimated system states are at least slightly deviated from the actual system states. On the other hand, since we train the agent (defender) with low-magnitude attacks with some known attack types (to create the effect of small deviations from actual system operation in the \emph{post-attack} state) and we test the proposed detector against various cyber-attacks in the numerical section (see Sec.~\ref{numerical}), we lay the main emphasis on online attack detection in this study. In general, the proposed detector can be considered as an online anomaly detection algorithm.}

{\textit{Remark 2:} The proposed solution scheme can be applied in a distributed smart grid system, where the learning and detection tasks are still performed in a single center but the meter measurements are obtained in a distributed manner. We briefly explain this setup as follows:
\begin{itemize}
  \item In the wide-area monitoring model of smart grids, there are several local control centers and a global control center. Each local center collects and processes measurements of a set of smart meters in its neighborhood, and communicates with the global center and with the neighboring local centers.
  \item The system state is estimated in a distributed manner, e.g., using the distributed Kalman filter designed for wide-area smart grids in \cite{Necip18}.
  \item Let $\mathbf{h}_k^T \in \mathbb{R}^{N}$ be the $k$th row of the measurement matrix, i.e., ${\mathbf{H}^T = [\mathbf{h}_1, \dots, \mathbf{h}_K]}$. Then, estimate of the negative log-scaled likelihood, $\eta_t$, can be written as follows (see \eqref{eq:eta_t}):
      \begin{gather} \label{eq:eta_t_v2}
        \eta_t = \sum_{k=1}^{K} (y_{k,t} - \mathbf{h}_k^T \hat{\mathbf{x}}_{t|t})^2.
      \end{gather}
      By employing the distributed Kalman filter, the local centers can estimate the system state at each time $t$. Then, they can compute the term $(y_{k,t} - \mathbf{h}_k^T \hat{\mathbf{x}}_{t|t})^2$ for the meters in their neighborhood. Let the number of local centers be $R$ and the set of meters in the neighborhood of the $r$th local center be denoted with $\mathcal{S}_r$. Then, $\eta_t$ in \eqref{eq:eta_t_v2} can be rewritten as follows:
      \begin{align} \nonumber
        \eta_t &= \sum_{r=1}^{R} \underbrace{\sum_{k \in \mathcal{S}_r} (y_{k,t} - \mathbf{h}_k^T \hat{\mathbf{x}}_{t|t})^2}_{\eta_{t,r}} \\ \nonumber
        &=  \sum_{r=1}^{R} \eta_{t,r}.
      \end{align}
  \item In the distributed implementation, each local center can compute $\eta_{t,r}$ and report it to the global center, which then sums $\{\eta_{t,r}, r = 1,2,\dots,R\}$ and obtain $\eta_t$.
  \item The learning and detection tasks (Algorithms 1 and 2) are then performed at the global center in the same way as explained above.
\end{itemize}}

\section{Simulation Results} \label{numerical}

\subsection{Simulation Setup and Parameters}

Simulations are performed on an IEEE-14 bus power system that consists of $N+1 = 14$ buses and $K = 23$ smart meters. The initial state variables (phase angles) are determined using the DC optimal power flow algorithm for case-14 in MATPOWER \cite{Zimmerman11}. The system matrix $\mathbf{A}$ is chosen to be an identity matrix and the measurement matrix $\mathbf{H}$ is determined based on the IEEE-14 power system. The noise variances for the normal system operation are chosen as $\sigma_v^2 = 10^{-4}$ and $\sigma_w^2 = 2 \times 10^{-4}$. 

For the proposed RL-based online attack detection scheme, the number of quantization levels is chosen as $I = 4$ and the quantization thresholds are chosen as $\beta_1 = 0.95\times10^{-2}$, $\beta_2 = 1.05\times10^{-2}$, and $\beta_3 = 1.15\times10^{-2}$ via an offline simulation by monitoring $\{\eta_t\}$ during the normal system operation. Further, $M = 4$ is chosen, i.e., sliding observation window consists of $4$ entries. Moreover, the learning parameters are chosen as $\alpha = 0.1$ and $\epsilon = 0.1$, and the episode length is chosen to be $T = 200$. In the learning phase, the defender is firstly trained over $4\times10^5$ episodes where the attack launch time is $\tau = 100$ and then trained further over $4\times10^5$ episodes where $\tau = 1$ to ensure that the defender sufficiently explores the observation space under normal operating conditions as well as the attacking conditions. More specifically, since a learning episode is terminated whenever the action \emph{stop} is chosen and observations under an attack become available to the defender only for $t \geq \tau$, we choose $\tau = 1$ in the half of the learning episodes to make sure that the defender is sufficiently trained under the post-attack regime.

To illustrate the tradeoff between the average detection delay and the false alarm probability, the proposed algorithm is trained for both $c = 0.02$ and $c = 0.2$. Moreover, to obtain a detector that is robust and effective against small deviations of measurements from the normal system operation, the defender needs to be trained with very low-magnitude attacks that correspond to slight deviations from the baseline. For this purpose, some known attack types with low magnitudes are used. In particular, in one half of the learning episodes, random FDI attacks are used with attack magnitudes being realizations of the uniform random variable $\pm \, \mathcal{U}[0.02,0.06]$, i.e., $b_{k,t} \sim \mathcal{U}[0.02,0.06]$, $\forall k \in \{1,\dots,K\}$, $\forall t \geq \tau$. In the other half of the learning episodes, random hybrid FDI/jamming attacks are used where $b_{k,t} \sim \mathcal{U}[0.02,0.06]$, $u_{k,t} \sim \mathcal{N}(0,\sigma_{k,t})$, and $\sigma_{k,t} \sim \mathcal{U}[2\times10^{-4},4\times10^{-4}]$, $\forall k \in \{1,\dots,K\}$, $\forall t \geq \tau$. 

\subsection{Performance Evaluation}

\begin{figure}[t]
\center
  \includegraphics[width=77mm]{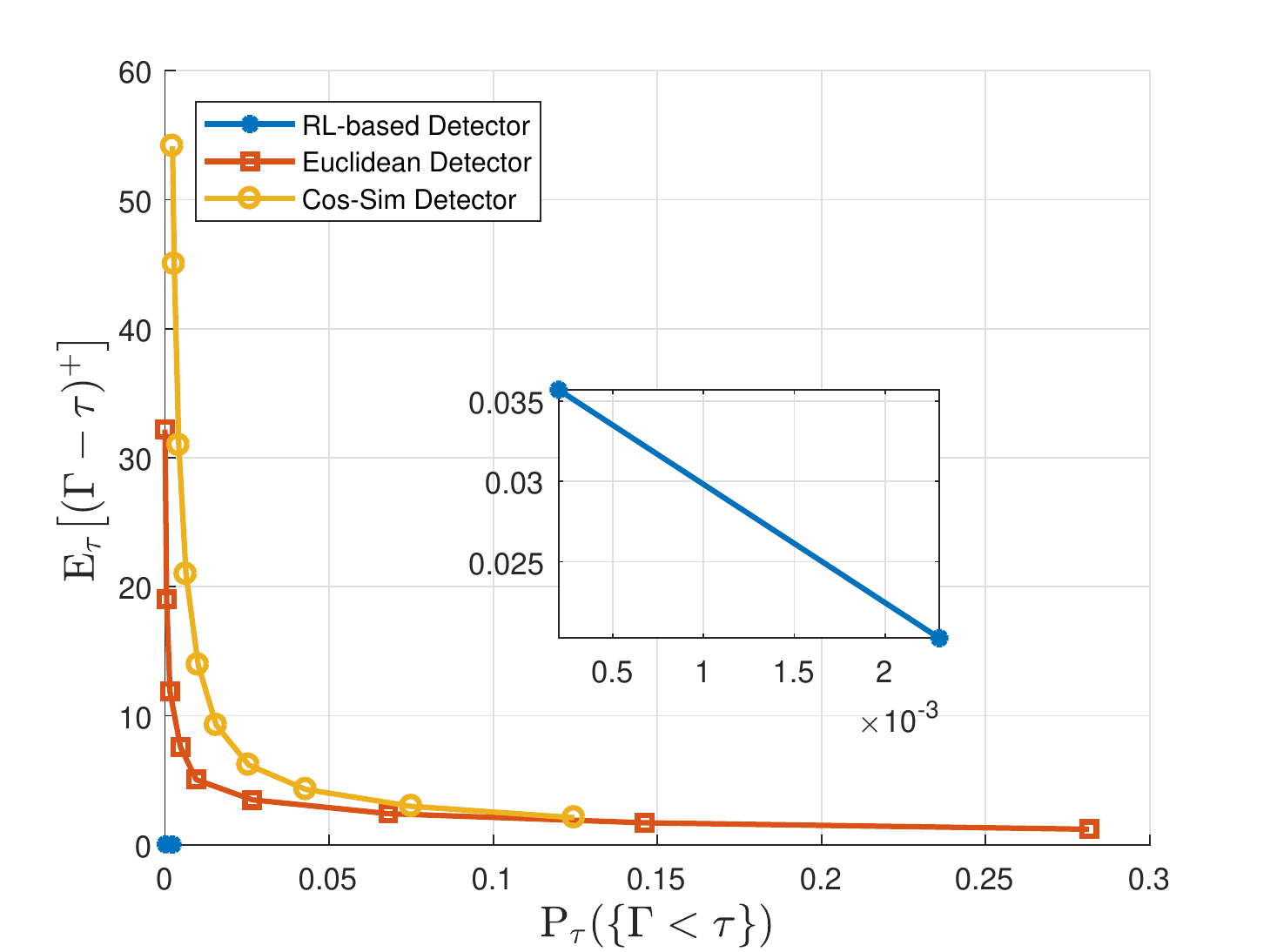}
\caption{Average detection delay vs. probability of false alarm curves for the proposed algorithm and the benchmark tests in case of a random FDI attack.}
 \label{fig:FDI}
\end{figure}

In this section, performance of the proposed RL-based attack detection scheme is evaluated and compared with some existing detectors in the literature. {Firstly, we report the average false alarm period, $\mathrm{E}_\infty[\Gamma]$, of the proposed detection scheme, i.e., the first time on the average the proposed detector gives an alarm although no attack/anomaly happens at all ($\tau = \infty$). The average false alarm periods are obtained as $\mathrm{E}_\infty[\Gamma] = 9.4696\times10^5$ for $c=0.2$ and $\mathrm{E}_\infty[\Gamma] = 7.9210\times10^6$ for $c=0.02$. As expected, false alarm rate of the proposed detector reduces as the relative cost of the false alarm event, $1/c$, increases.}

Based on the optimization problem in \eqref{eq:opt_prob1}, our performance metrics are the probability of false alarm, i.e., $\mathrm{P}_\tau (\{\Gamma < \tau\})$, and the average detection delay, i.e., $\mathrm{E}_\tau \big[(\Gamma-\tau)^+\big]$. Notice that both performance metrics depend on the unknown attack launch time $\tau$. Hence, in general, the performance metrics need to be computed for each possible $\tau$. For a representative performance illustration, we choose $\tau$ as a geometric random variable with parameter $\rho$ such that $P(\tau = k) = \rho \, (1-\rho)^{k-1}, k = 1,2,3, \dots$ where $\rho \sim \mathcal{U}[10^{-4},10^{-3}]$ is a uniform random variable.

With Monte Carlo simulations over 10000 trials, we compute the probability of false alarm and the average detection delay of the proposed detector, the Euclidean detector \cite{Manandhar14}, and the cosine-similarity metric based detector \cite{Rawat15}. To obtain the performance curves, we vary the thresholds of the benchmark tests and vary $c$ for the proposed algorithm. To evaluate the proposed algorithm, we use Algorithm~\ref{alg:test} that makes use of the $Q$-tables learned in Algorithm~\ref{alg:training} for $c = 0.02$ and $c = 0.2$. {Furthermore, we report the precision, recall, and F-score for all simulation cases. As the computation of these measures requires computing the number of detected and missed trials, we define an upper bound on the detection delay (that corresponds to the maximum acceptable detection delay) such that if the attack is detected within this bound we assume the attack is detected, otherwise missed. As an example, we choose this bound as $10$ time units. Then, we compute the precision, recall, and F-score out of $10000$ trials as follows:
\begin{equation}\nonumber
\mbox{Precision} = \frac{\mbox{\# trials } (\tau \leq \Gamma \leq \tau + 10)}{\mbox{\# trials } (\tau \leq \Gamma \leq \tau + 10) + \mbox{\# trials } (\Gamma < \tau)},
\end{equation}
\begin{equation}\nonumber
\mbox{Recall} = \frac{\mbox{\# trials } (\tau \leq \Gamma \leq \tau + 10)}{\mbox{\# trials } (\tau \leq \Gamma \leq \tau + 10) + \mbox{\# trials } (\Gamma > \tau + 10)},
\end{equation}
and
\begin{equation}\nonumber
\mbox{F-score} = 2 ~ \frac{\mbox{Precision} \times \mbox{Recall}}{\mbox{Precision} + \mbox{Recall}},
\end{equation}
where ``\# trials'' means ``the number of trials with''.}

\begin{figure}[t]
\center
  \includegraphics[width=77mm]{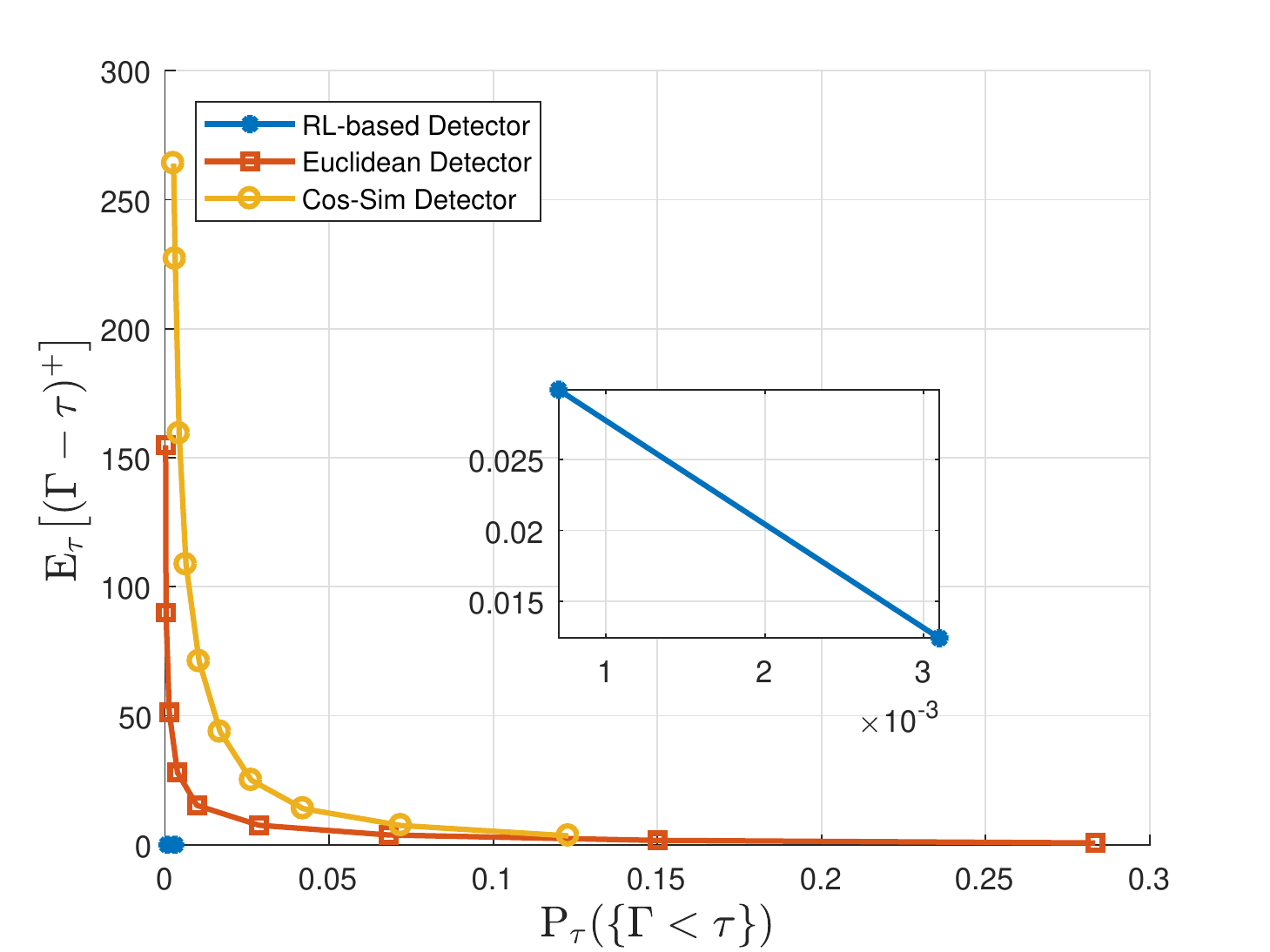}
\caption{{Performance curves for the proposed algorithm and the benchmark tests in case of a structured ``stealth'' FDI attack.}}
 \label{fig:stealth_FDI}
\end{figure}

\begin{figure}[t]
\center
  \includegraphics[width=77mm]{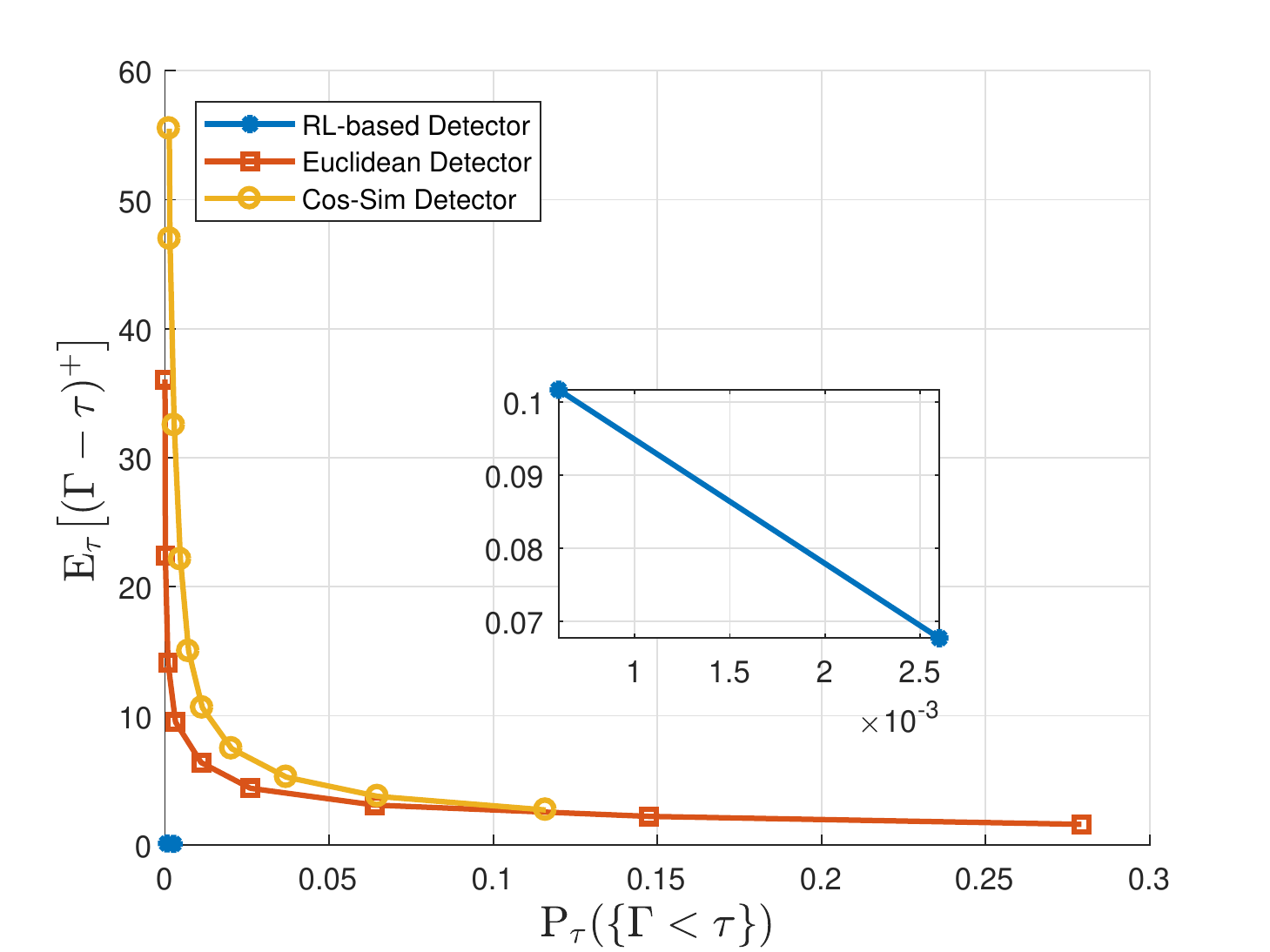}
\caption{Performance curves for the proposed algorithm and the benchmark tests in case of a jamming attack with AWGN.}
 \label{fig:jamming}
\end{figure}

\begin{figure}[t]
\center
  \includegraphics[width=77mm]{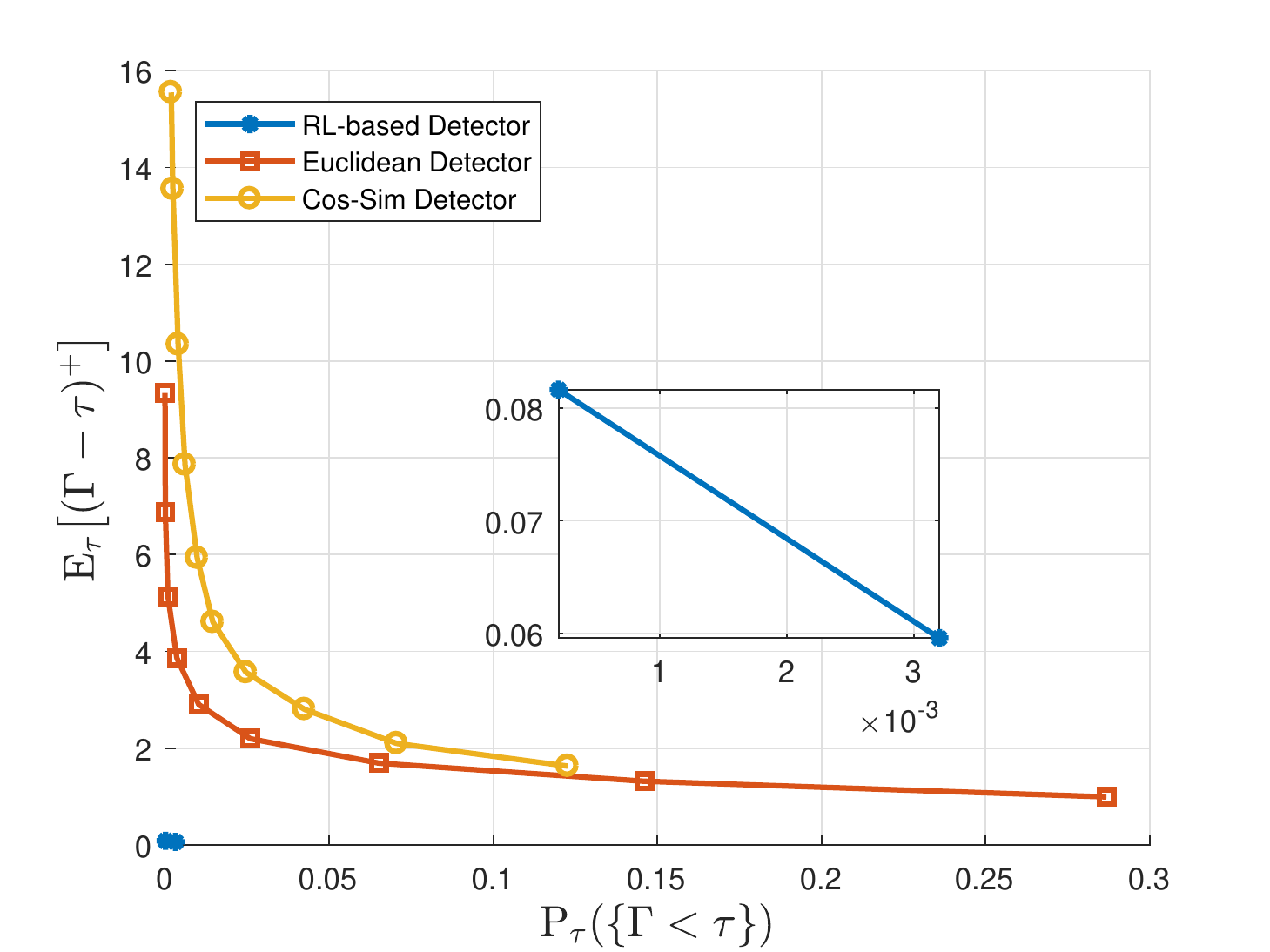}
\caption{Performance curves for the proposed algorithm and the benchmark tests in case of a jamming attack with jamming noise correlated over the space.}
 \label{fig:corr_jamm}
\end{figure}

\begin{figure}[t]
\center
  \includegraphics[width=77mm]{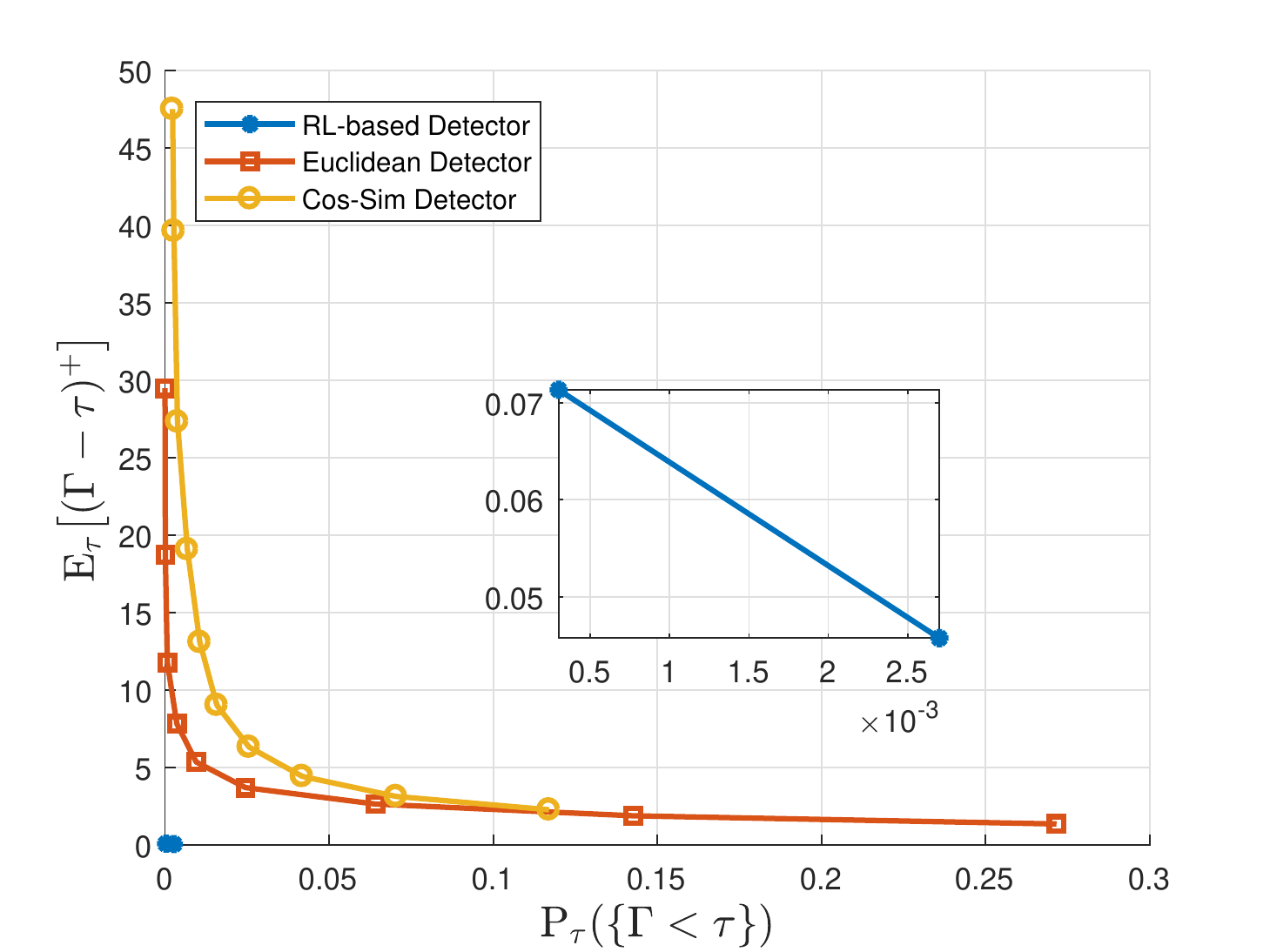}
\caption{Performance curves for the proposed algorithm and the benchmark tests in case of a hybrid FDI/jamming attack.}
 \label{fig:hybrid}
\end{figure}

\begin{figure}[t]
\center
  \includegraphics[width=77mm]{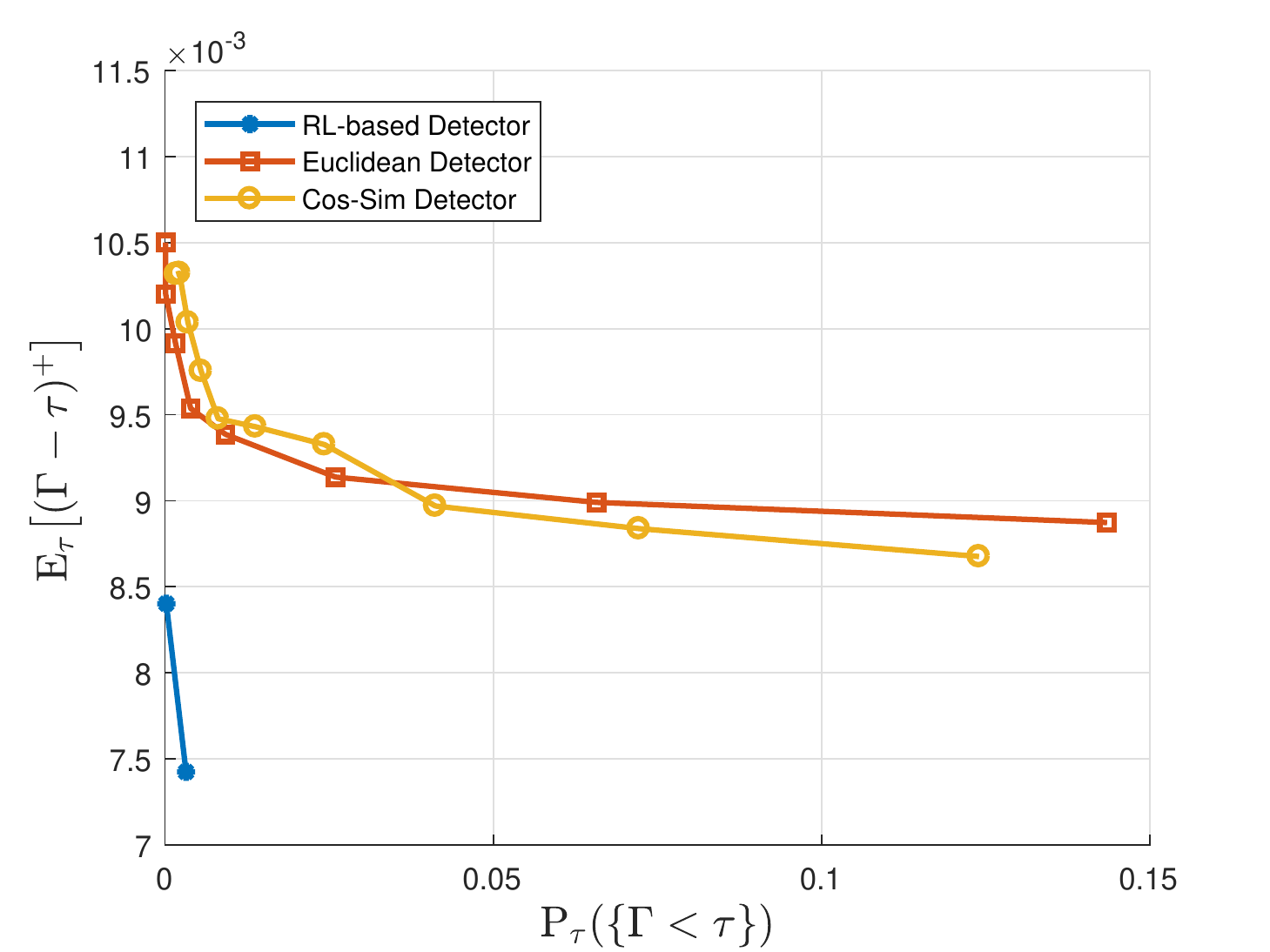}
\caption{Performance curves for the proposed algorithm and the benchmark tests in case of a DoS attack.}
 \label{fig:dos}
\end{figure}

\begin{figure}[t]
\center
  \includegraphics[width=77mm]{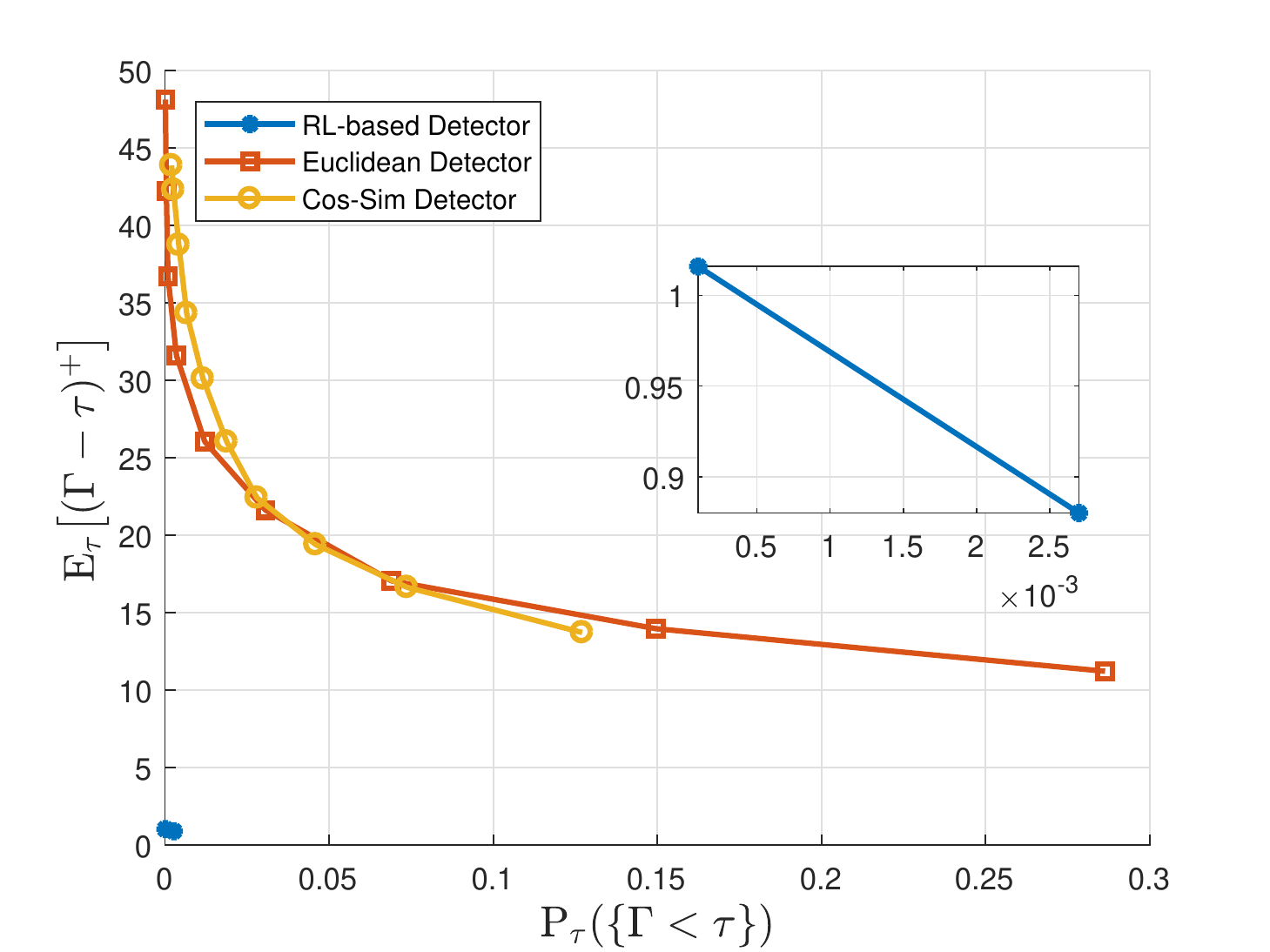}
\caption{{Performance curves for the proposed algorithm and the benchmark tests in case of a network topology attack.}}
 \label{fig:topology}
\end{figure}

\begin{figure}[t]
\center
  \includegraphics[width=77mm]{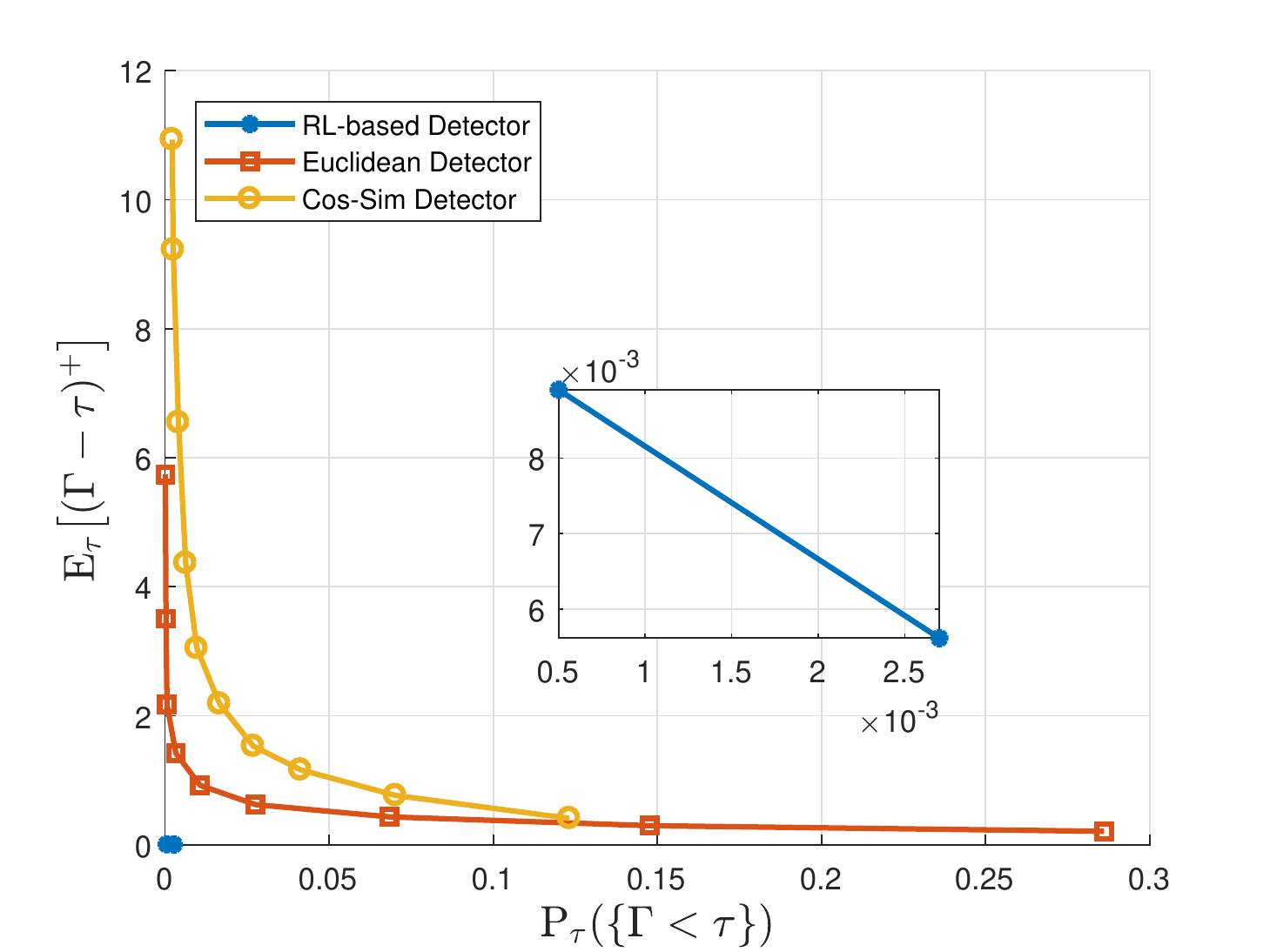}
\caption{{Performance curves for the proposed algorithm and the benchmark tests in case of a mixed network topology and hybrid FDI/jamming attack.}}
 \label{fig:mixed}
\end{figure}

\begin{figure}[t]
\center
  \includegraphics[width=77mm]{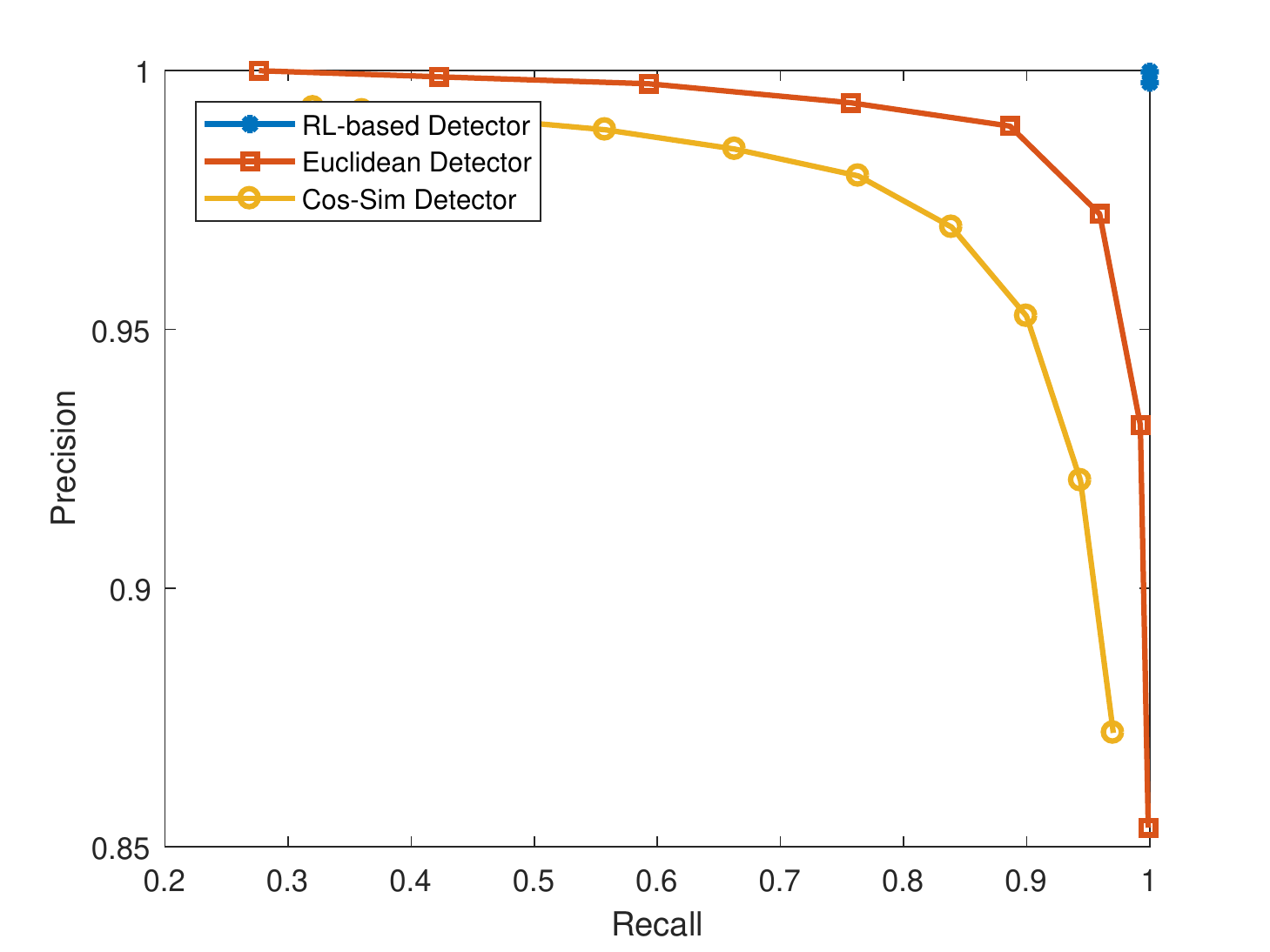}
\caption{{Precision vs. recall for the proposed and the benchmark detectors against a random FDI attack.}}
 \label{fig:F-measures}
\end{figure}

We evaluate the proposed and the benchmark detectors under the following attack scenarios:
\begin{enumerate}
  \item Firstly, we evaluate the detectors against a random FDI attack {where $b_{k,t} \sim \mathcal{U}[-0.07,0.07]$, $\forall k \in \{1,\dots,K\}$ and $\forall t \geq \tau$.} The corresponding tradeoff curves are presented in Fig.~\ref{fig:FDI}.
  \item {We then evaluate the detectors against a structured ``stealth'' FDI attack \cite{Liu09}, where the injected data $\mathbf{b}_t$ lies on the column space of the measurement matrix $\mathbf{H}$. We choose $\mathbf{b}_t = \mathbf{H} \mathbf{g}_t$ where $\mathbf{g}_t \triangleq [g_{1,t}, \dots, g_{N,t}]^\mathrm{T}$ and $g_{n,t} \sim \mathcal{U}[0.08,0.12]$, $\forall n \in \{1,\dots,N\}$ and $\forall t \geq \tau$. The corresponding performance curves are illustrated in Fig.~\ref{fig:stealth_FDI}.}
  \item Then, we evaluate the detectors in case of a jamming attack with zero-mean AWGN where {$u_{k,t} \sim \mathcal{N}(0,\sigma_{k,t})$ and $\sigma_{k,t} \sim \mathcal{U}[10^{-3},2\times10^{-3}]$, $\forall k \in \{1,\dots,K\}$ and $\forall t \geq \tau$.} The corresponding tradeoff curves are presented in Fig.~\ref{fig:jamming}.
  \item Next, we evaluate the detectors in case of a jamming attack with jamming noise correlated over the meters where $\mathbf{u}_t \sim  \mathbf{\mathcal{N}}(\mathbf{0},\mathbf{U}_t)$, $\mathbf{U}_t = \pmb{\Sigma}_t \pmb{\Sigma}_t^\mathrm{T}$, and $\pmb{\Sigma}_t$ is a random Gaussian matrix with its entry at the $i$th row and the $j$th column is $\pmb{\Sigma}_{t,i,j} \sim \mathcal{N}(0,8\times10^{-5})$. The corresponding performance curves are given in Fig.~\ref{fig:corr_jamm}.
  \item Moreover, we evaluate the detectors under a hybrid FDI/jamming attack {where $b_{k,t} \sim \mathcal{U}[-0.05,0.05]$, $u_{k,t} \sim \mathcal{N}(0,\sigma_{k,t})$, and $\sigma_{k,t} \sim \mathcal{U}[5\times10^{-4},10^{-3}]$, $\forall k \in \{1,\dots,K\}$ and $\forall t \geq \tau$.} The corresponding tradeoff curves are presented in Fig.~\ref{fig:hybrid}.
  \item Then, we evaluate the detectors in case of a random DoS attack where the measurement of each smart meter become unavailable to the system controller at each time with probability $0.2$. That is, for each meter $k$, $d_{k,t}$ is $0$ with probability $0.2$ and $1$ with probability $0.8$ at each time $t \geq \tau$. The performance curves against the DoS attack are presented in Fig.~\ref{fig:dos}.
   \item {Further, we consider a network topology attack where the lines between the buses 9-10 and 12-13 break down. The measurement matrix $\bar{\mathbf{H}}_t$ for $t\geq\tau$ is changed accordingly. The corresponding tradeoff curves are given in Fig.~\ref{fig:topology}.}
   \item {Finally, we consider a mixed topology and hybrid FDI/jamming attack, where the lines between buses 9-10 and 12-13 break down for $t\geq\tau$ and further, we have $b_{k,t} \sim \mathcal{U}[-0.05,0.05]$, $u_{k,t} \sim \mathcal{N}(0,\sigma_{k,t})$, and $\sigma_{k,t} \sim \mathcal{U}[5\times10^{-4},10^{-3}]$, $\forall k \in \{1,\dots,K\}$ and $\forall t \geq \tau$. The corresponding performance curves are presented in Fig.~\ref{fig:mixed}.}
\end{enumerate}
{Table~\ref{table:performance_c0p2} and Table~\ref{table:performance_c0p02} summarize the precision, recall, and F-score for the proposed RL-based detector for $c=0.2$ and $c=0.02$, respectively against all the considered simulation cases above. Moreover, for the random FDI attack case, Fig.~\ref{fig:F-measures} illustrates the precision versus recall curves for the proposed and benchmark detectors. Since we obtain similar results for the other attack cases, we report the results for the random FDI attack case as a representative.}

For almost all cases, we observe that the proposed RL-based detection scheme significantly outperforms the benchmark tests. This is because through the training process, the defender learns to differentiate the instantaneous high-level system noise from persistent attacks launched to the system. Then, the trained defender is able to significantly reduce its false alarm rate. Moreover, since the defender is trained with low attack magnitudes, it becomes sensitive to detect small deviations of the system from its normal operation. On the other hand, the benchmark tests are essentially outlier detection schemes making sample-by-sample decisions and hence they are unable to distinguish high-level noise realizations from real attacks that makes such schemes more vulnerable to false alarms. Finally, in case of DoS attacks, since the meter measurements become partially unavailable so that the system greatly deviates from its normal operation, all detectors are able to detect the DoS attacks with almost zero average detection delays (see Fig.~\ref{fig:dos}).


\renewcommand{\arraystretch}{1.1}
\begin{table*}[t]

    \centering
    \begin{tabular}{ | p{1.6cm} | l | l | l | l | l | l | l | l |}
    \hline
    Measure & FDI & Jamming & Corr. Jamm. & Hybrid & DoS & Structured FDI & Topology & Mixed \\ \hline \hline
    Precision & 0.9977 & 0.9974 & 0.9968 & 0.9973 & 0.9977 & 0.9968 & 0.9972 & 0.9973   \\ \hline
    Recall & 1 & 1 & 1 & 1 & 1 & 0.9756 & 0.9808 & 1 \\ \hline
    F-score & 0.9988 & 0.9987 & 0.9984 & 0.9986 & 0.9988 & 0.9861 & 0.9890 & 0.9986 \\
    \hline
    \end{tabular}
    \vspace{-0.05cm}
    \caption{{Precision, recall, and F-score for the proposed detector ($c = 0.2$) in detection of various cyber-attacks.}}
    \label{table:performance_c0p2}
\end{table*}

\begin{table*}[t]

    \centering
    \begin{tabular}{ | p{1.6cm} | l | l | l | l | l | l | l | l |}
    \hline
    Measure & FDI & Jamming & Corr. Jamm. & Hybrid & DoS & Structured FDI & Topology & Mixed \\ \hline \hline
    Precision  & 0.9998 & 0.9994 & 0.9998 & 0.9997 & 0.9995 & 0.9993 & 0.9999 & 0.9995   \\ \hline
    Recall & 1 & 1 & 1 & 1 & 1 & 0.9449 & 0.9785 & 1 \\ \hline
    F-score & 0.9999 & 0.9997 & 0.9999 & 0.9998 & 0.9997 & 0.9713 & 0.9891 & 0.9997 \\
    \hline
    \end{tabular}
    \vspace{-0.05cm}
    \caption{{Precision, recall, and F-score for the proposed detector ($c = 0.02$) in detection of various cyber-attacks.}}
    \label{table:performance_c0p02}
\end{table*}
\renewcommand{\arraystretch}{1}

%

\section{Concluding Remarks} \label{conclusion}

In this paper, an online cyber-attack detection problem is formulated as a POMDP problem and a solution based on the model-free RL for POMDPs is proposed. The numerical studies illustrate the advantages of the proposed detection scheme in fast and reliable detection of cyber-attacks targeting the smart grid. The results also demonstrate the high potential of RL algorithms in solving complex cyber-security problems. In fact, the algorithm proposed in this paper can be further improved using more advanced methods. Particularly, the following directions can be considered as future works:
\begin{itemize}
  \item compared to the finite-size sliding window approach, more sophisticated memory techniques can be developed,
  \item compared to discretizing the continuous observation space and using a tabular approach to compute the $Q$ values, linear/nonlinear function approximation techniques, e.g., neural networks, can be used to compute the $Q$ values,
  \item and deep RL algorithms can be useful to improve the performance.
\end{itemize}

Finally, we note that the proposed online detection method is widely applicable to any quickest change detection problem where the pre-change model can be derived with some accuracy but the post-change model is unknown. This is, in fact, commonly encountered in many practical applications where the normal system operation can be modeled sufficiently accurately and the objective is the online detection of anomalies/attacks that are difficult to model. Moreover, depending on specific applications, if real post-change, e.g., attack/anomaly, data can be obtained, the real data can be further enhanced with simulated data and the training can be performed accordingly, that would potentially improve the detection performance.



\bibliography{Refs_SaTC17,det_refs}
\bibliographystyle{IEEEtran}

\end{document}